\documentclass[runningheads]{llncs}
\pdfoutput=1
\usepackage{graphicx}
\usepackage{comment}
\usepackage{amsmath,amssymb} 
\usepackage{color}
\usepackage{csquotes}
\usepackage{import}
\usepackage{multirow}

\usepackage{booktabs,array}
\usepackage{enumitem}
\usepackage{amssymb}
\usepackage{authblk}

\usepackage{tablefootnote}
\usepackage{xspace}
\graphicspath{ {figures/} }

\newcommand{\etal}{\textit{et al}.}

\makeatletter
\renewcommand\AB@affilsepx{, \protect\Affilfont}
\makeatother

\begin{document}
\pagestyle{headings}
\mainmatter
\def\ECCVSubNumber{7036}  

\title{$S^3$Net: Semantic-Aware Self-supervised \\Depth Estimation with Monocular Videos and Synthetic Data} 

\titlerunning{$S^3$Net: Semantic-Aware Self-supervised Depth Estimation}
%
\author{Bin Cheng\inst{1} \and
Inderjot Singh Saggu \inst{2,5} \and
Raunak Shah \inst{3} \and
Gaurav Bansal\inst{4} \and
Dinesh Bharadia\inst{5}
}



%
\authorrunning{B. Cheng et al.}

\institute{
Rutgers University, NJ, USA \and
Plus.ai, Cupertino, USA~\footnote{work done at UCSD} \and
Indian Institute of Technology, Kanpur, India$^{**}$ \and
Blue River Technology, Sunnyvale, USA \and
University of California, San Diego, USA \\
\email{cb3974@winlab.rutgers.edu, inderjot.saggu@plus.ai, raunaks@iitk.ac.in, gaurav.bansal@bluerivert.com, dineshb@ucsd.edu}}
\maketitle

\begin{abstract}
Solving depth estimation with monocular cameras enables the possibility of widespread use of cameras as low-cost depth estimation sensors in applications such as autonomous driving and robotics. 
However, learning such a scalable depth estimation model would require a lot of labeled data which is expensive to collect. There are two popular existing approaches which do not require annotated depth maps:
(i) using labeled synthetic and unlabeled real data in an adversarial framework to predict more accurate depth, and
(ii) unsupervised models which exploit geometric structure across space and time in monocular video frames.
Ideally, we would like to leverage features provided by both approaches as they complement each other; however, existing methods do not adequately exploit these additive benefits.
We present $S^3$Net, a self-supervised framework which combines these complementary features: we use synthetic and real-world images for training while exploiting geometric, temporal, as well as semantic constraints. Our novel consolidated architecture provides a new state-of-the-art in self-supervised depth estimation using monocular videos. We present a unique way to train this self-supervised framework, and achieve (i) more than $15\%$ improvement over previous synthetic supervised approaches that use domain adaptation and (ii) more than $10\%$ improvement over previous self-supervised approaches which exploit geometric constraints from the real data.

\end{abstract}

\section{Introduction}
Depth estimation is a fundamental component of 3D scene understanding, with applications in fields such as autonomous driving, robotics and space exploration. There has been considerable progress in estimating depth through monocular camera images in the last few years, as monocular cameras are inexpensive and widely deployed on many robots. However, building supervised depth estimation algorithms using monocular cameras is challenging, primarily because collecting ground-truth depth maps for training requires a carefully calibrated setup. As an example, many vehicles currently sold in the market have monocular cameras deployed, but there is no trivial way to obtain ground-truth depth information from the images collected from these cameras. Thus, supervised methods for depth estimation suffer due to the unavailability of extensive training labels. 

To overcome the lack of depth annotation for monocular camera data, existing work has explored two areas of research: either designing self-supervised/semi-supervised approaches which require minimal labeling, or leveraging labeled synthetic data. Most self-supervised approaches rely on geometric and spatial constraints~\cite{geonet2018}, and have succeeded in reducing the impact of this issue, however they don't always perform well in challenging environments with conditions like limited visibility, object motion, etc. This is because they lack strong training signal from supervision which lets them learn from and generalize to these conditions. In contrast, some effort has been undertaken to use realistic simulated environments to obtain additional synthetic depth data which can be used to compute a supervised training loss. 

Synthetic data can be easily generated in different settings with depth labels - for example by varying the lighting conditions, changing the weather, varying object motion, etc. Simply training the original model on synthetic data, however, does not work well in practice as the model does not generalize well to a real-world dataset. To bridge this domain gap between the real-world and synthetic datasets, many domain adaptation techniques have been proposed. Recent works, like~\cite{t2net2018,mou2019learning}, have found success in using adversarial approaches to address this issue. These solutions typically involve using an adversarial transformation network to align the domains of the synthetic and real-world images, followed by a task network that is responsible for predicting depth. Naturally, we pose the question - can we build a depth estimation network that combines the benefits of information conveyed through real as well as synthetic data?

\begin{figure}[!t] 
\label{fig:overview}
    \centering\includegraphics[width=0.9\textwidth]{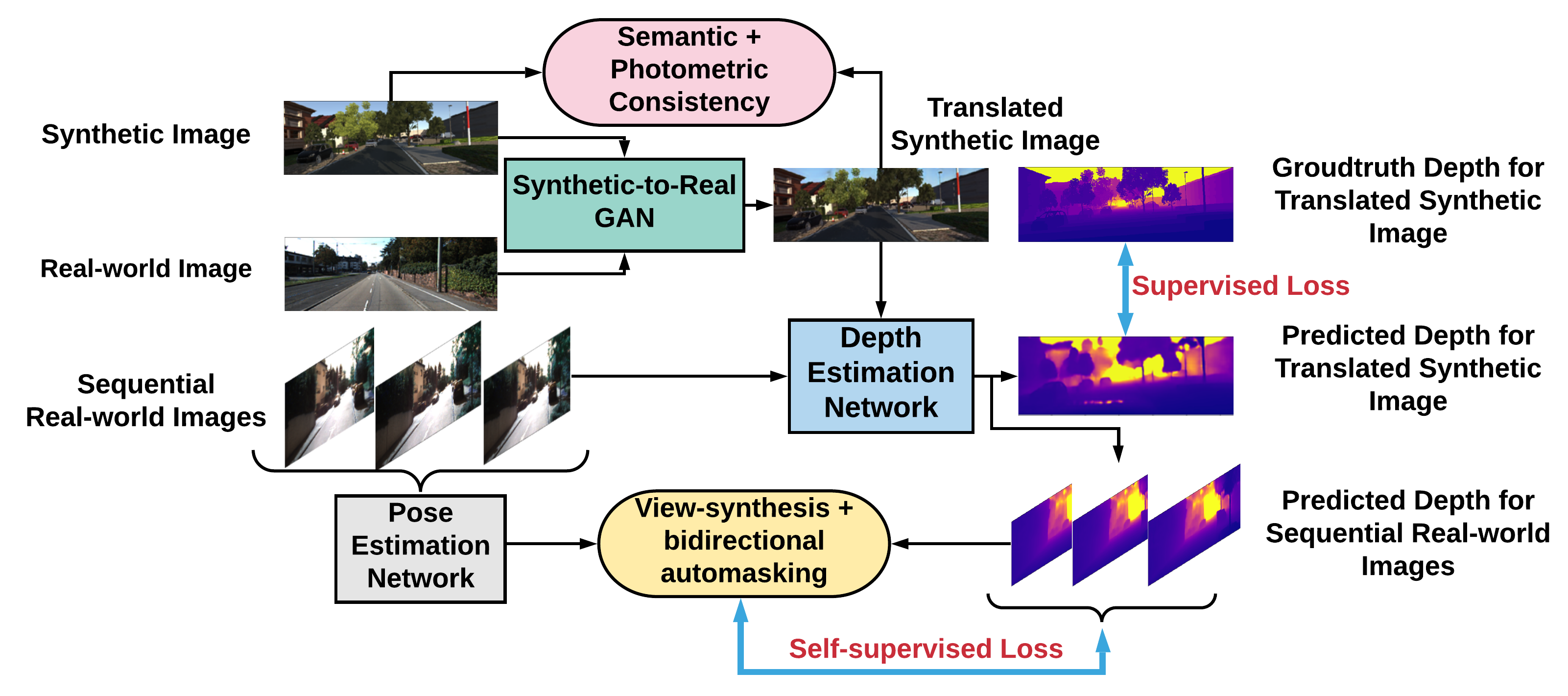}
    \caption{Overview of our proposed framework: integrating supervised learning on translated synthetic data and self-supervised learning on real videos while imposing spatial, temporal and semantic constraints}
\end{figure}

We present a novel framework $S^3$Net that trains the depth network by exploiting these self-supervised constraints (derived from real-world sequential images) and supervised constraints (derived from synthetic data and the respective ground-truth depths). This framework is implemented through several integrated stages which are described below:
First, as shown in Fig.~\ref{fig:overview}, we present a novel Generative Adversarial Network (GAN)-based domain adaptation network which exploits geometric constraints across space and time, as well as the semantic consistency between original synthetic images and translated images. These constraints encode additional latent information and thus enhance the quality of domain adaptation. Next, to leverage \enquote*{synthetic} supervised cues and \enquote*{real} self-supervised cues, we present a novel training approach - weights of the depth estimation network are updated \textit{alternatively} based on the supervised and self-supervised losses. Finally, to impose explicit constraints on object geometry we augment the input RGB images with semantic labels, and utilize a bi-directional auto-masking technique to limit pixels which would violate rigid motion constraints.

\textbf{Novel Adversarial Framework:} 
The key idea of our GAN structure is to utilize flow-based photometric consistency and semantic consistency to better guide the image translation and reduce the domain gap. By utilizing the flow and the sequential translated images, the frame at $t$ can be used to reconstruct the frame at $t + 1$. The photometric differences between the reconstructed frame $t+1$ and the original frame $t+1$ are primarily due to imperfect image translation. Moreover, the semantic information should remain consistent before and after the image translation. Therefore, we add both a photometric consistency and a semantic consistency loss to create a novel adversarial framework. These offer additional constraints on the domain adaptation and further improve the image translation performance. They also help increase robustness and reduce undesired artifacts in translated images when compared with traditional approaches, as described in Section 4.2.

\textbf{Semantics and Bi-directional Auto-Masking:}
Inspired by the auto-masking technique proposed in~\cite{godard2018digging}, we propose a novel bi-directional auto-masking technique for sequential real-world images, which can filter out the pixels violating the fundamental rigid motion assumption for self-supervised depth learning. The key difference from a single direction mask is that the bi-directional technique fuses the masks learned by reconstructing frame $t + 1$ from frame $t$ and vice versa, which can substantially increases the accuracy of the proposed mask. Moreover, we augment the input images of our model with semantic labels. The semantic labels can provide explicit geometry constraints, which can be beneficial to further boost the performance of the image translation and the depth estimation. 

The challenges of training our depth model fall under two major categories: the GAN networks are unstable during training due to the presence of supervised synthetic losses and self-supervised losses, which results in lack of convergence. We address the convergence issue by proposing a two-phase training strategy. In the first phase, we train the image translation network and depth estimation network with synthetic supervised losses to stabilize the GAN-based image translation network. In the second phase, we freeze the weights of the image translation network and further train the depth estimation network with both supervised and self-supervised losses. 

We evaluate our framework on two challenging datasets, i.e., KITTI~\cite{kitti2015} and Make3D~\cite{saxena2008make3d}. The evaluation results show that our proposed model can outperform the state-of-the-art approaches in all evaluated metrics. In particular, we show that $S^3$Net can outperform both the state-of-the-art synthetic supervised domain adaptation approaches~\cite{mou2019learning} by $\sim{15}\%$ and self-supervised approaches~\cite{godard2018digging} by $\sim{10}\%$. Moreover, we only require the depth estimation network during inference, so our inference compute requirements are comparable to previous state-of-the-art approaches.

\section{Related Work}
Monocular depth estimation is often considered an ill-posed problem, since a single 2D image can be produced from an infinite number of distinct 3D scenes. Without additional constraints, it is challenging to predict depths correctly for a given image. Previous works address this challenge in two ways: (i) supervised depth estimation trained with ground-truth depths; (ii) self-supervised depth estimation that learns indirect depth cues from sequential images. 

\subsection{Supervised Depth Estimation}
Eigen~\etal~\cite{eigen2014depth} proposed the first supervised learning architecture that models multi-scale information for pixel-level depth estimation using direct regression. Inspired by this work, many follow-up works have extended supervised depth estimation in various directions~\cite{he2018learning,repala2018dual,liu2016learning,xu2018structured,cao2017estimating,roy2016monocular,chen2016single-image,zhang2015monocular,xu2017multiscale,xu2018attention,fu2018deep,xian2018monocular}. However, acquiring these ground-truth depths is prohibitively expensive. Therefore, it is unlikely to obtain a large amount of labelled training data covering various road conditions, weather conditions, etc, which indicates that these approaches may not generalize well.

One promising approach that reduces the labeling cost is to use synthetically generated data. However, models trained on synthetic data typically perform poorly on real-world images due to a large domain gap. Domain adaptation aims to minimize this gap. 
Recently, GAN-based approaches show promising performance in domain adaptation~\cite{ganin2014unsupervised,ganin2016domain,sankaranarayanan2018generate,bousmalis2017unsupervised,Sankaranarayanan_2018_CVPR}. 
Atapour~\etal~\cite{atapour2018real} proposed a CycleGAN-based translator~\cite{zhu2017unpaired} to translate real-world images into the synthetic domain, and then train a depth prediction network using the synthetic labeled data.
Zheng~\etal~\cite{zheng2018t2net} propose a novel architecture ($T^2$Net) where the style translator and the depth estimation network are optimized jointly so that they can improve each other. Despite promising performance, these approaches inherently suffer from mode collapse and semantic distortion due to imperfect synthetic-to-real image translation. Various constraints and techniques have been proposed to improve the quality of the translated images, but image translation~\footnote{``domain adaptation" and ``image translation" are used interchangeably} still remains a challenging task.

\subsection{Self-supervised Depth Estimation}
In addition to supervised solutions, various approaches have been studied to predict depths by extracting disparity and depth cues from stereo image pairs or monocular videos. Garg~\etal~\cite{garg2016unsupervised} introduced a warping loss based on Taylor expansion. An image reconstruction loss with a spatial smoothness constraint was introduced in~\cite{ren2017unsupervised,zhou2017unsupervised,jason2016back} to learn depth and camera motion. Recent works~\cite{vijayanarasimhan2017sfmnet,sfm-learner,mahjourian2018unsupervised,godard2017unsupervised,godard2018digging} aim to improve depth estimation by further exploiting geometry constraints. In particular, Godard~\etal~\cite{godard2017unsupervised} employed epipolar geometry constraints between stereo image pairs and enforced a left-right consistency constraint in training the network. Zhou~\etal~\cite{zhou2017unsupervised} proposed a network to learn pose and depth from videos by introducing a photometric consistency loss while only relying on monocular videos for training. Yin~\etal\cite{geonet2018} proposed GeoNet, which also used depth and pose networks in order to compute rigid flow between sequential images in a video. More specifically, they introduced a temporal, flow-based photometric loss to predict depth for monocular videos in an unsupervised setting. 
Bian~\etal~\cite{bian2019unsupervised} used a similar approach along with a self discovered mask to handle dynamic and occluded objects. Gordon~\etal~\cite{gordon2019depth} also addresses these issues in a purely geometric approach. Casser~\etal~\cite{casser2018depth} adapts a similar framework with an additional online refinement model during inference. Xu~\etal~\cite{xu2019region} uses region deformer networks along with the earlier constraints to handle rigid and non-rigid motion. Zhou~\etal~\cite{zhou2019unsupervised} use a dual network attention based model which processes low and high resolution images separately. Godard~\etal~\cite{godard2018digging} also presented another unsupervised approach which built on their earlier model~\cite{godard2017unsupervised} by modifying the implementation of the unsupervised constraints. 

Another set of recently adopted approaches involves using semantic information, which provides additional constraints on object geometry that can potentially boost the accuracy of depth estimation~\cite{signet2019,ramirez2018geometry,chen2019towards,ranjan2018competitive}.  Meng~\etal~\cite{signet2019} built on top of~\cite{geonet2018}, and proposed several ways to implement a semantic aided network which helped improve performance.  
Ranjan~\etal~\cite{ranjan2018competitive} used a competitive collaboration framework to leverage segmentation maps, pose and flow for depth estimation. However, even with various constraints, the self-supervised approaches predict depth primarily based on indirect and weak-supervision depth cues, which can be easily affected by undesired artifacts, such as motion blurring and low visibility. 

Our model architecture is influenced by various previous work, e.g. approaches in~\cite{godard2018digging,zheng2018t2net,mou2019learning,semgan}. But, compared to these approaches, our $S^3$Net cooperatively combines both supervised depth prediction on synthetic data and self-supervised depth prediction on sequential images, such that the two strategies can complement each other in a mutually beneficial setting.

\section{Proposed Methods}
We propose a joint framework for monocular depth estimation that is trained on translated synthetic images in a supervised manner and further fine-tuned on sequences of real-world images in a self-supervised fashion. Our proposed framework can be broken down into two main components: 
a)  Synthetic-to-Real Translation and Task Prediction ($G_{S\rightarrow R}$, $D_R$, $f_T$), and,  b) View-synthesis guided self-supervised fine-tuning (\textit{Pose}, $f_T$)

\begin{figure}[htb]
    \centering\includegraphics[width=\textwidth]{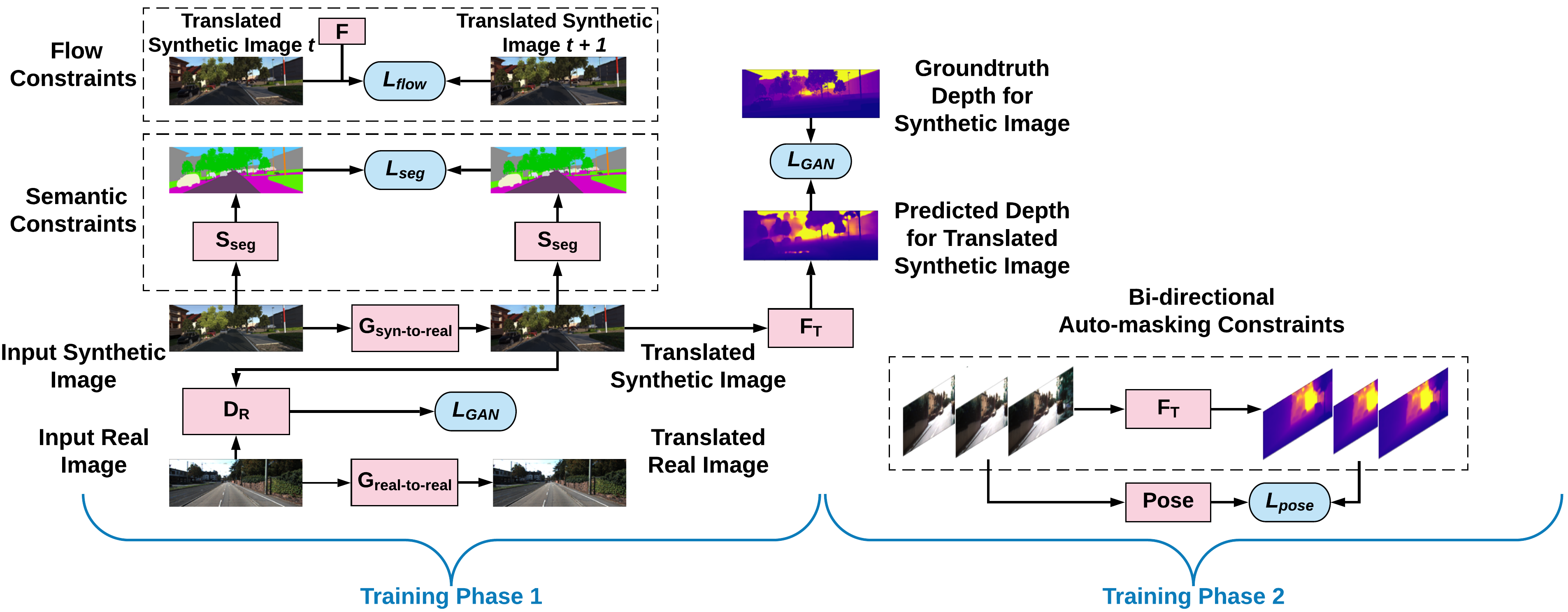}
    \caption{Our detailed  architecture, a) Semantic and photometric consistent GAN for synthetic supervised depth estimation, b) self-supervised architecture trained on sequence of real-world images with warping view-synthesis loss.}\label{fig:architecture}
\end{figure}

\subsection{Novel GAN Architecture}
Models trained on synthetic data do not generalize well to real-world data because of domain shift. To address this problem, we build upon the work of $T^2$Net ~\cite{t2net2018} for supervised depth estimation on translated synthetic images. 

\subsubsection{Adversarial Constraints}
The goal of our generator is to translate a synthetic image $X_s$ to the real domain $X_r$. To achieve this, 
a discriminator $D_R$ and a transformer architecture $G_{S\rightarrow R}$  are trained jointly such that discriminator tries to predict if the image is real or synthetic. This accounts for our GAN loss, $\mathcal{L}_{GAN}$ as shown in Fig.~\ref{fig:architecture}: 

\begin{equation}
    \mathcal{L}_{GAN} = \mathbb{E}_{x_r \sim X_r }[\log D_R(x_r)] + \mathbb{E}_{x_s \sim X_s }[\log(1- D_R(G_{S\rightarrow R}(x_s) ) ) ]
\end{equation}

\subsubsection{Identity Constraints}
To improve the quality of translated images, $T^2$Net imposes a identity constraint such that if a real image $x_r$ is given as input, the generator network ($G_{S\rightarrow R}$)'s output should be the identical real image $x_r$. This additional constraint is incorporated as an identity loss $\mathcal{L}_r$  in Fig. \ref{fig:architecture}: 
\begin{equation}
    \mathcal{L}_{r} = ||G_{S\rightarrow R}(x_r) - x_r ||_{1}
\end{equation}

\subsubsection{Semantic Consistency}
While the identity constraint improves upon a vanilla GAN architecture, we observe that the translated images had artifacts as shown in Fig. \ref{fig:translated_images}, which cause imperfect domain translation and subsequently hurt depth prediction. 
To address this we introduce a semantic consistency loss $\mathcal{L}_{seg}$  (Fig. \ref{fig:architecture}) based on the idea that given a semantic segmentation model $S_{seg}$ trained on the source domain, $x_s$ and $G_{S\rightarrow R }(x_s)$ should have identical semantic segmentation maps. This is intuitive as domain translation shouldn't affect the semantic structure of the image. We enforce this by treating $S_{seg}(x_s)$ as a ground truth label for pixel-wise prediction scores $S_{seg}(G_{S\rightarrow R }(x_s))$. These are used to compute a cross-entropy loss function over semantic labels: 
\begin{equation}
    \mathcal{L}_{seg} = -\sum_{pixels}\sum_{labels} S_{seg}(x_s)\log (S_{seg}(G_{S\rightarrow R }(x_s)))
\end{equation}

However, because of domain shift we cannot expect $S_{seg}$ trained on synthetic images to generalize well to the translated image domain, and hence we also continue training $S_{seg}$ while training our GAN architecture so that it can learn features that are generalized to both domains.

\subsubsection{Photometric Consistency with Ground-truth Flow}
In addition to semantic constraints we introduce a flow guided photometric loss\cite{geonet2018} to exploit the temporal structure in translated image sequences.
By applying ground-truth flow, a frame $t$ can be used to reconstructed the frame $t+1$. We represent this as a transformation $\mathcal{F}$. In Eq.~\ref{eqn:synthetic_photometric_loss} below, $G_{S\rightarrow R}(x_{s,t})$ represents the translated image of a synthetic frame at time $t$ and $\mathcal{F}(G_{S\rightarrow R}(x_{s,t}))$ indicates the reconstructed frame $t+1$ based on frame $t$. $L_{pe}(*)$ computes the photometric differences between the reconstructed frame and the true frame. This photometric loss provides an indirect supervision on the synthetic-to-real image translation.

\begin{equation} \label{eqn:synthetic_photometric_loss}
    \mathcal{L}_{flow} = L_{pe}(\mathcal{F}(G_{S\rightarrow R}(x_{s,t})), G_{S\rightarrow R}(x_{s,t+1})) 
\end{equation}

Incorporating the above constraints in our GAN framework results in an improved domain translator that is largely devoid of artifacts and preserves semantic structure as shown in Fig. \ref{fig:translated_images}.

\subsection{Combining Supervised and Self-supervised Depth Estimation}
\label{sec_3.2}
\subsubsection{Supervised Depth Estimation on Synthetic Data}
With the ground-truth depth labels for synthetic data, we formulate the depth estimation on synthetic data as a regression problem. 
In Eq.~\ref{eqn:task} below, $f_T(G_{S\rightarrow R}(x_{s,t}))$ is the estimated depth map for frame $t$ and $y_{s, t}$ is the correspond ground-truth label for synthetic frame $t$. 
\begin{equation} \label{eqn:task}
    \mathcal{L}_{task} = ||f_T(G_{S\rightarrow R}(x_{s,t})) - y_{s, t}||_1
\end{equation}
In accordance with our base network ($T^2$Net) we add an edge consistency/awareness loss which penalizes discontinuity (or inconsistency) in the edges between the image $x_s$ and its depth map $f_T(x_s)$. \begin{equation}
    \mathcal{L}_s = |\partial_xf_T(x_r)|e^{-|\partial_xx_r|} + |\partial_yf_T(x_r)|e^{-|\partial_yx_r|}
\end{equation}
Our training is divided into two phases. In the first phase, we train a GAN-based image transfer network (($G_{S\rightarrow R}$ and $D_R$)) and a depth estimation network ($f_{T}$). A detailed explanation of our training methodology follows in Section $3.4$. The first loss objective is a weighted combination of the above constraints (Sections 3.1, 3.2):
\begin{equation} \label{eqn:supervised}
    \mathcal{L}_{phase1} = \mathcal{L}_{GAN} + \alpha_{r} \mathcal{L}_r + \alpha_{seg} \mathcal{L}_{seg} + \alpha_{flow} \mathcal{L}_{flow} + \alpha_{task} \mathcal{L}_{task} + \alpha_{s} \mathcal{L}_s
\end{equation}
where $\alpha_{r}$, $\alpha_{seg}$, $\alpha_{flow}$, $\alpha_{task}$, and $\alpha_{s}$ are hyper-parameters.

\subsubsection{Self-supervised Depth Estimation on Monocular Videos}
In addition to supervised depth prediction on translated synthetic images, we also perform self-supervised depth estimation on monocular videos. The corresponding pixel coordinates of one rigid object in two consecutive frames follows the relationship 
\begin{align}
    \mathrm{p}_{t+1} = \mathbf{K}\mathbf{T}_{t\rightarrow t+1}\mathbf{f}_{T}(\mathrm{p}_t)\mathbf{K}^{-1}\mathrm{p}_t
    && \mathrm{p}_{t+1} = \mathcal{W}^{-1} \mathrm{p}_{t}
\end{align}
where $\mathrm{p}_{t}$ and $\mathrm{p}_{t+1}$ are the positions of a pixel in frames at time $t$ and $t+1$,  $\mathbf{K}$ denotes the camera intrinsic parameters, $\mathbf{T}_{t\rightarrow t+1}$ represents the relative camera pose from frame $t$ to from $t + 1$ and $\mathcal{W}$ is the equivalent warping transformation. By sampling these pixels at $\mathrm{p}_{t+1}$ in frame $t+1$, one can construct frame $t$ from frame $t+1$. The photometric difference (denoted by $L_{pe}$) between the constructed image and the true image of frame $t$ provides self-supervision for both the depth network ($f_{T}$) and pose estimation networks (Fig. \ref{fig:architecture}).

\begin{equation} \label{}
    \mathcal{L}_{pose} = L_{pe}(\mathcal{W}(x_{r,t+1}),\: x_{r, t})
\end{equation}

\subsubsection{Bi-directional Auto-Masking}
Inspired by the auto-masking method proposed in~\cite{godard2018digging}, we compute the photometric loss for different sequential image pairs, e.g., from frame $t + 1$ to frame $t$ and from frame $t - 1$ to frame $t$, and then aggregate these photometric losses by extracting their minimum value, i.e.,  $\min \limits_{t' \in \{\dots t-1, t+1 \dots\}} L_{pe}(\mathcal{W}(x_{r,t'}),\: x_{r, t})$. Additionally, the pixels satisfying 

\[mask = \min \limits_{t' \in \{\dots t-1, t+1 \dots\}} L_{pe}(x_{r,t'},\: x_{r, t}) > \min \limits_{t' \in \{\dots t-1, t+1 \dots\}} L_{pe}(\mathcal{W}(x_{r,t'}),\: x_{r, t})\] 
are selected for further loss computation. It is because the discarded pixels are more likely to belong to moving objects with a similar moving speed as the moving camera, or stationary objects captured by a stationary camera. For a more complete loss computation, we consider a bi-directional warping transformation, i.e., from frame $t$ to frame $t'$ as well as from frame $t'$ to frame $t$.

\begin{equation} \label{}
\begin{split}
    \mathcal{L}_{mask} = & mask_{t'\rightarrow t} \circ \min \limits_{t' \in \{\dots t-1, t+1 \dots\}} L_{pe}(\mathcal{W}(x_{r,t'}),\: x_{r, t}) \\ 
    & + mask_{t\rightarrow t'} \circ \min \limits_{t' \in \{\dots t-1, t+1 \dots\}} L_{pe}(\mathcal{W}(x_{r,t}),\: x_{r, t'}))
\end{split}
\end{equation}
In the second phase of our training, we train the depth and pose networks ($f_{T}$ and $Pose$) with a combination of sequential real images and GAN translated synthetic images. Our total loss objective which includes training the depth network $f_{T}$ with supervised loss $\mathcal{L}_{task}$ from synthetic data, can be written as:
\begin{equation}
    \mathcal{L}_{phase2} = \alpha_{pose} \mathcal{L}_{pose} + \alpha_{mask} \mathcal{L}_{mask} + \alpha_{task} \mathcal{L}_{task} 
\end{equation}
where $\alpha_{pose}$, $\alpha_{mask}$ and $\alpha_{task}$ are hyper-parameters. More details regarding our training strategy follow in Section $3.4$.

\subsection{Semantic Augmentation}
Semantic labels provide important information about object shape and geometry. We believe such information helps improve the accuracy of depth estimation by imposing additional constraints. For example, on 2D images, the pixels on the object boundaries can have very different depths. Semantic information can help regulate the pixels belonging to certain objects and facilitate the learning process of depth estimation. In this work, to utilize semantic information, we augment the input RGB images with additional semantic labels. We also experimented with augmenting RGB images with semantic labels during synthetic-to-real image translation and obtained substantial improvements in the quality of our translated images. We did not apply the semantic consistency loss defined in Section 3.1 while conducting these experiments. Results for this study are provided in Table \ref{tab:ablation_study}.

\section{Experiments}
We first present the implementation details of our model, including the network implementation, data pre-processing, and our training and inference strategies. We test our model on the KITTI and Make3D benchmarks and compare our performance with other state-of-the-art models. Finally, we study the importance of each component in our model through various ablation experiments.

\subsection{Implementation Details}
\subsubsection{Network Implementation}
Our framework mainly consists of two main sub-modules: (i) the syn-to-real image translation network which translates synthetic images to real-style images, and (ii) the depth estimation network which predicts depth maps for both translated synthetic images and real videos.
For the synthetic-to-real image translation networks, we build our network on the $T^2$Net architecture~\cite{t2net2018}, with added constraints that use the synthetic ground-truth labels for semantic and optical flow. For the semantic consistency loss we use DeepLab v3+ with a MobileNet backbone as our $S_{seg}$ model. We pre-trained the model on vKITTI, achieving a median IoU of 0.898 on the validation set. 
We tested the U-Net, VGGNet, and ResNet50 architectures for the depth estimation network and selected the U-Net architecture due to its best performance.
A VGG-based architecture was used to estimate the relative camera poses between sequential images. These depth maps and camera poses are subsequently used to compute the self-supervised loss. 

\subsubsection{Data Pre-processing}
We use vKITTI~\cite{gaidon2016vkitti} and KITTI~\cite{kitti2015} as the synthetic dataset and the real-world dataset, respectively while training the synthetic-to-real image translation network. The training dataset consists of 20470 images from vKITTI and 41740 images from KITTI.
The training images of the KITTI dataset are further divided into small sequences. Images in each sequence are ordered so that they represent a short video clip. We use 697 images from KITTI as our test dataset as per the eigen split~\cite{eigen2014depth}. The input images are resized to $640\times192$ (width $\times$ height) during both training and testing. The ground-truth depth information, semantic labels, and optical flow information from the synthetic vKITTI dataset are also used during training. As discussed in \cite{t2net2018}, the maximum vKITTI ground-truth depth is 655.3m, whereas the maximum KITTI depth is about 80m. Thus, we clip the synthetic ground truth depths to the range of [0.01, 80] meters. For our real data, we require semantic labels in addition to the monocular video images from the KITTI dataset. We use the pre-trained DeepLab v3+ model~\cite{deeplabv3plus2018} to generate semantic labels for real images.

\subsubsection{Model Training}
\label{sec:training_strategy}
Training our model has two major challenges: (i) the training of GAN-based networks is known to be unstable; (ii) the depth estimation in our model consists of two components and thus the weights of the depth estimation network are updated by two separate loss functions, which can lead to a convergence issue. To tackle this problem, we design a two-phase training strategy. In the first phase, we pre-train the synthetic-to-real image translation network along with synthetic supervised depth estimation constraints to provide a stable initialization for the GAN-based image translation network. In the second phase, we freeze the weights of the image translation network and train the depth estimation network using both supervised and self-supervised losses. We primarily tested two training methods to harmonize the two sources of losses: 1) \textit{weighted sum training}: updating the weights of the depth estimation network based on a weighted sum of the two sources of losses; 2) \textit{alternating training}: alternatively updating the weights of the depth estimation networks by the two sources of losses. 
We find that the two training methods are resulting in comparable evaluation results but the alternating training provides another control knob to optimize the model training, and generalize well to both the data sources and therefore to unseen datasets. Due to space limitation, we show the results using alternating training only in this paper. 

Further, we use the Adam optimizer~\cite{kingma2014adam}, with initial learning rate of $2e^{-5}$ for the image translation network, $5e^{-5}$ for the depth estimation network, and $5e^{-5}$ for the camera pose estimation network. 

Our network was trained on a RTX 2080Ti GPU and the training took 2.6 hours per epoch. On average, our depth estimation network can process 33 frames per second during inference.

\subsection{Monocular Depth Estimation on KITTI Dataset}
We follow the procedure defined in~\cite{t2net2018} when evaluating on the KITTI dataset. First, the ground truth depths are generated by projecting 3D LiDAR points to the image plane and then the depth predictions are clipped at a distance of 80m and 50m. The evaluation results are listed in Table~\ref{tab:results_comparison}, where all metrics are computed according to the evaluation strategy proposed in~\cite{eigen2014depth}. 
As shown in the table, our model has the best performance across all metrics. We believe this is because our model can synergize the merits of supervised depth estimation with domain adaptation and self-supervised depth estimation with real-world images. Typical supervised synthetic approaches train models by using low-cost synthetic ground-truth depth, but these approaches also suffer from unstable and inconsistent image translation, leading to less accurate translated images with low resolution. On the other hand, self-supervised approaches can learn the depth from high resolution sequential images; however, these depths are learned from indirect cues which are sensitive to in-view object movements, blockages, etc. Training the model with modified supervised and self-supervised constraints in our consolidated framework ensures that we exploit the best of both worlds, which ultimately leads to better prediction results. 

\begin{table}[hbt]
\small
\begin{center}
\resizebox{\textwidth}{!}
{
\begin{tabular}{c ||c| c|c|c|c | c|c|c}
    \hline
    \multirow{2}{*}{Method} &  \multirow{2}{*}{Dataset} 
    & \multicolumn{4}{c|}{Error-related metrics} & \multicolumn{3}{c}{Accuracy-related metrics} \\
    & & \textbf{Abs Rel} & \textbf{Sq Rel} & \textbf{RMSE} & \textbf{RMSE log} 
    & $\mathbf{\delta < 1.25}$ & $\mathbf{\delta < 1.25^{2}}$  & $\mathbf{\delta < 1.25^{3}}$ \\
    \hline
    \multicolumn{1}{c}{} & \multicolumn{1}{c}{} & \multicolumn{4}{c}{\textbf{depth capped at 80m}} & \multicolumn{3}{c}{} \\
    \hline
    \hline
    Zhou~\etal~\cite{zhou2017unsupervised} & K & 0.183 & 1.595 & 6.709 & 0.270 & 0.734 & 0.902 & 0.959 \\
    \hline
    Yin~\etal~\cite{geonet2018} & K & 0.155 & 1.296 & 5.857 & 0.233 & 0.793 & 0.931 & 0.973 \\
    \hline
    Wang~\etal~\cite{ddvowang2018cvpr} & K & 0.151 & 1.257 & 5.583 & 0.228 & 0.810 & 0.936 & 0.974 \\ 
    \hline
    Ramirez~\etal~\cite{ramirez2018geometry} & K  & 0.143 & 2.161 & 6.526 & 0.222 & 0.850 & 0.939 & 0.972 \\
    \hline
    Casser~\etal~\cite{casser2018depth} & K & 0.141 & 1.026 & 5.290 & 0.215 & 0.816 & 0.945 & 0.979 \\
    \hline
    Ranjan~\etal~\cite{ranjan2018competitive} & K & 0.140 & 1.070 & 5.326 & 0.217 & 0.826 & 0.941 & 0.975 \\
    \hline
    Xu~\etal~\cite{xu2019region} & K & 0.138 & 1.016 & 5.352 & 0.216 & 0.823 & 0.943 & 0.976 \\
    \hline
    Meng~\etal~\cite{signet2019} & K & 0.133 & 0.905 & 5.181 & 0.208 & 0.825 & 0.947 & 0.981 \\
    \hline
    Godard~\etal~\cite{godard2018digging}~\tablefootnote{For fair comparison, we selected the results for the model without pre-training on the ImageNet dataset}
    & K & 0.132 & 1.044 & 5.142 & 0.210 & 0.845 & 0.948 & 0.977 \\ 
    \hline
    Zheng~\etal~\cite{t2net2018} & K + V & 0.174 & 1.410 & 6.046 & 0.253 & 0.754 & 0.916 & 0.966 \\
    \hline
    Mou~\etal~\cite{mou2019learning} & K + V & 0.145 & 1.058 & 5.291 & 0.215 & 0.816 & 0.941 & 0.977\\ 
    \hline
    Ours & K + V & \textbf{0.124} & \textbf{0.826} & \textbf{4.981} & \textbf{0.200} & \textbf{0.846} & \textbf{0.955} & \textbf{0.982} \\
    \hline
    \hline
    \multicolumn{1}{c}{} & \multicolumn{1}{c}{} & \multicolumn{4}{c}{\textbf{depth capped at 50m}} & \multicolumn{3}{c}{} \\
    \hline
    \hline
    Yin~\etal~\cite{geonet2018}  & K & 0.147 & 0.936 & 4.348 & 0.218 & 0.810 & 0.941 & 0.977 \\
    \hline
    Zheng~\etal~\cite{t2net2018} & K +V & 0.168 & 1.199 & 4.674 & 0.243 & 0.772 & 0.912 & 0.966 \\
    \hline
    Mou~\etal~\cite{mou2019learning} & K + V & 0.139 & 0.814 & 3.995 & 0.203 & 0.830 & 0.949 & 0.980\\ 
    \hline
    Ours & K + V & \textbf{0.118} & \textbf{0.615} & \textbf{3.710} & \textbf{0.187} & \textbf{0.862} & \textbf{0.962}	& \textbf{0.984} \\
    \hline
\end{tabular}
}
\end{center}

\caption{Monocular depth estimation on KITTI dataset with Eigen~\etal~\cite{eigen2014depth} split. The highlighted scores mark the best performance among selected models. In ``Dataset" column, ``K" and ``V" stands for the KITTI and the vKITTI dataset, respectively.}\label{tab:results_comparison}
\end{table}

\begin{figure}[hbt] 
    \centering\includegraphics[width=0.9\textwidth]{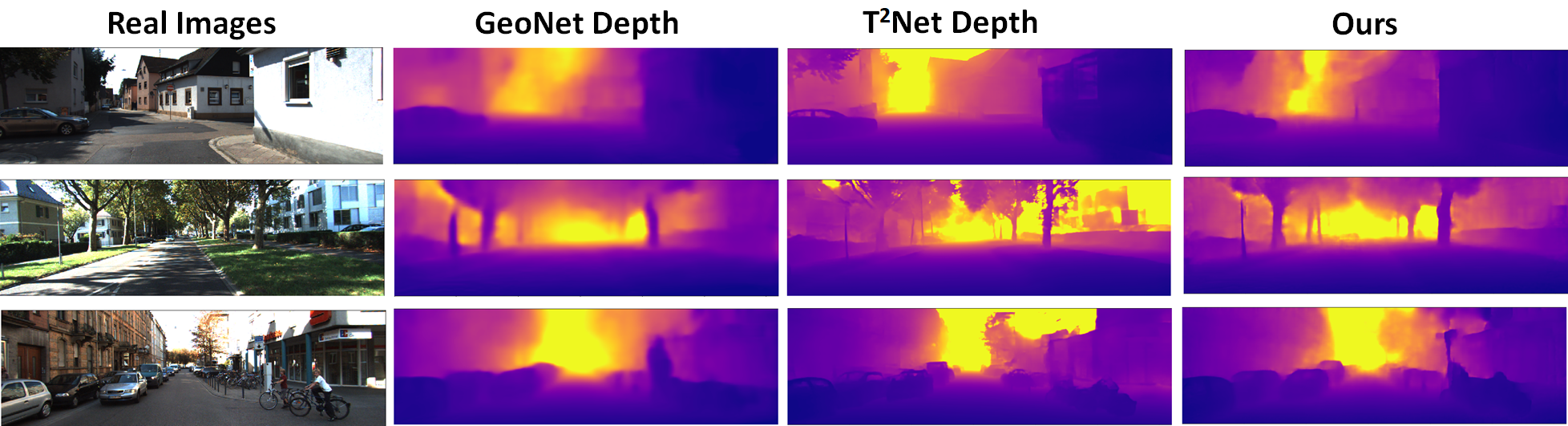}
    \caption{Qualitative Depth Prediction Results: Column (a) real-world images from KITTI, Column (b), (c), (d) are results for  GeoNet~\cite{geonet2018}, $T^2$Net~\cite{t2net2018}, and our $S^3$Net framework, respectively.} \label{fig:depth}
\end{figure}

In Fig.~\ref{fig:depth} we compare qualitative depth estimation results of purely self-supervised GeoNet~\cite{geonet2018}, purely synthetic supervised $T^2$Net~\cite{t2net2018} and our proposed framework. Purely self-supervised approaches results in depth maps which are blurred and do not model depth discontinuity at object boundaries well. On the other hand purely synthetic supervised approach results in sharper depth maps but because of imperfect domain translation it fails to predict depth for surfaces with multiple textures. For example, in the first row of Fig~\ref{fig:depth}, $T^2$Net predicts incorrect depth values for the wall on the right because of the window on the wall adding additional texture. 
These defects severely limit the real-world application of purely self-supervised and synthetic supervised techniques. Our $S^3$Net on the other hand generates sharper depth maps than GeoNet and doesn't suffer from the problems discussed for $T^2$Net depth, further proving our point about combining best features from both. 

\begin{figure}[hbt] 
    \centering\includegraphics[width=0.9\textwidth]{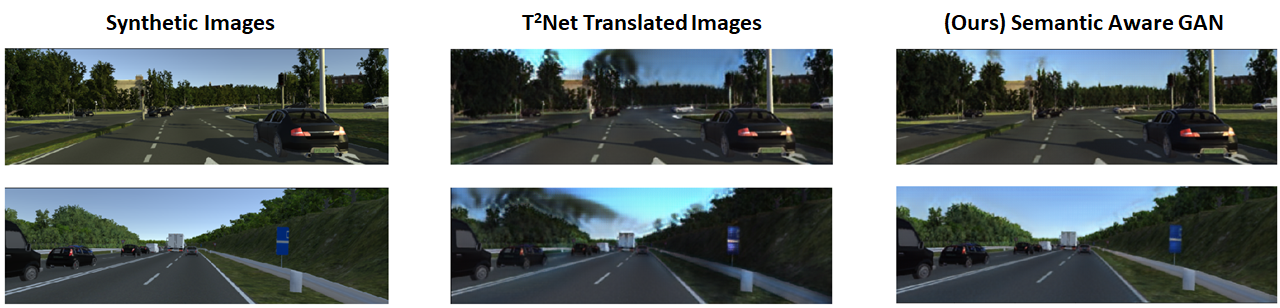}
    \caption{Translated images: Column (a) input synthetic images; Column (b) $T^2$Net; Column (c) our $S^3$Net GAN with semantic constraints.} \label{fig:translated_images}
\end{figure}

In Fig.\ref{fig:translated_images} we compare syn-to-real translated images for $T^2$Net GAN and our semantic consistent GAN. Without the presence of a specific task loss, e.g. a depth estimation loss, the resulting objective drives the image translator to generate a realistic interpretation of synthetic images. However, when a task loss is introduced the main objective is shifted to project synthetic images to a space that is optimized for the task. Therefore, some of these differences might not be visually perceivable but can lead to a large gain in performance. Our approach significantly reduces artifacts and successfully retains the semantic structure across synthetic and translated images.

\subsubsection{Camera Pose Estimation}
We train and evaluate our model's performance on the KITTI odometry dataset. Our training follows the same strategy as in Section~\ref{sec:training_strategy} - the only difference is in the real dataset we use. We train on sequences [``00", ``01", ... , ``08"], and test on sequences ``09" and ``10" as per the KITTI odometry split. We use a sequence length of 3 with the same architecture that was used in other experiments. Our evaluation follows the same strategy as given in~\cite{zhou2017unsupervised}. As shown in Table~\ref{tab:pose}, our model outperforms the state-of-the-art approaches by a convincing margin. 
\begin{table}[hbt]
\begin{center}
\resizebox{0.95\textwidth}{!}
{
\begin{tabular}{c|c|c|c}
\hline
Method & \textbf{\# of snippets} & \textbf{Seq.09} & \textbf{Seq.10} \\
\hline
ORB-SLAM~(full) & 5 & $0.014\pm 0.008$ & $0.012\pm 0.011$ \\\hline
ORB-SLAM~(short) & 5 & $0.064\pm 0.141$ & $0.064\pm 0.130$ \\\hline
DDVO (Wang~\etal~\cite{wang2018learning}) & 3 & $0.045 \pm 0.108$ & $0.033 \pm 0.074$ \\\hline
SfmLearner (Zhou~\etal~\cite{zhou2017unsupervised}) & 5 & $0.021\pm 0.017$ & $0.020\pm 0.015$ \\\hline
SfmLearner~\cite{zhou2017unsupervised} updated & 5 & $0.016\pm 0.009$ & $0.013\pm 0.009$ \\\hline
GeoNet (Yin~\etal~\cite{geonet2018}) & 5 &  $0.012 \pm 0.007$ & $0.012 \pm 0.009$\\ \hline
MonoDepth2* (Godard~\etal~\cite{godard2018digging}) & 2 & $0.017 \pm 0.008$ & $0.015 \pm 0.010$ \\\hline
EPC++ (Luo~\etal~\cite{luo2018every}) & 3 & $0.013 \pm 0.007$ & $0.012 \pm 0.008$ \\\hline
Ours & 3 & {\bf{ 0.0097 $\pm$ 0.0046 }}& \bf{0.0099 $\pm$ 0.0071} \\
\hline
\end{tabular}
}
\end{center}
\caption{Absolute Trajectory Error (ATE) on the KITTI odometry dataset.} 

\label{tab:pose}
\vspace{-2ex}
\end{table}

\subsection{Generalization Study on Make3D Dataset}
To show the generalization capability, we also test our framework on the Make3D dataset~\cite{saxena2008make3d}. We use the model trained using the KITTI dateset and the vKITTI dataset, and evaluate the model on the Make3D test dataset, following the evaluation strategy in~\cite{godard2017unsupervised}. 
As shown in Table~\ref{tab:make3d_results}, our $S^3$Net performs better than all other existing self-supervised monocular approaches. 

\begin{table}[hbt]
\begin{center}
\resizebox{0.95\textwidth}{!}
{
\begin{tabular}{c || c|| c|c|c|c}
    \hline
    \multirow{2}{*}{Method}
    & \multirow{2}{*}{Train}
    & \multicolumn{4}{c}{Error-related metrics} \\ 
    & & \textbf{Abs Rel} & \textbf{Sq Rel} & \textbf{RMSE} & \textbf{RMSE log} \\
    \hline
MonoDepth (Godard~\etal~\cite{godard2017unsupervised}) & No & 0.544 & 10.940 & 11.760 & 0.193
 \\ \hline
SfmLearner (Zhou~\etal~\cite{zhou2017unsupervised}) & No & 0.383 & 5.321 & 10.47 & 0.478
 \\ \hline
$T^2$Net (Zheng~\etal~\cite{t2net2018}) & No & 0.508 & 6.589 & 8.935 & 0.574
 \\ \hline
 MonoDepth2 (Godard~\etal~\cite{godard2018digging}) & No & 0.322 & 3.589 & 7.417 & 0.163
 \\ \hline
TCDA (Mou~\etal~\cite{mou2019learning}) & No & 0.384 & 3.885 & 7.645 & 0.181
 \\ \hline
Ours (no semantic augmentation) & No & 0.372 & 5.699 & 7.844 & 0.176
 \\ \hline
Ours (with semantic augmentation) & No & \textbf{0.322} & \textbf{3.238} & \textbf{7.187} & \textbf{0.164}
 \\ \hline
\end{tabular}
}
\end{center}
\caption{Error metrics for depth estimation on the Make3D dataset. ``Train" refers to whether the model was trained on the Make3D train set.}\label{tab:make3d_results}
\end{table}

\subsection{Ablation Study}
In this subsection, we perform a set of ablation experiments on the KITTI dataset to discuss how each individual component in our framework contributes to the final performance. The evaluation results are reported in Table~\ref{tab:ablation_study}.

\subsubsection{Synthetic Translated Supervised Depth Estimation}
Due to a large domain gap between the synthetic and the real domain, a model that is only trained on synthetic data typically generates unacceptable depth predictions when tested on real data. Synthetic-to-real image translation is one of the most effective remedies for this issue. Even with a native image translation network as proposed in~\cite{t2net2018}, the depth predictions on the real data can be improved by about 40\%. Additionally, the flow-guided photometric consistency and semantic consistency constraint further regulates the image translation and improves our depth prediction accuracy by another 8.4\% and 8\% respectively. Continuing training $S_{seg}$ gives better performance compared to freezing network parameters. This is because $S_{seg}$ is trained on synthetic semantic labels and cannot generalize to translated domain if not trained further.

\subsubsection{Synthetic Translated Supervised + Semantic Augmentation}
We investigate the importance of semantic augmentation to our model by (i) only augmenting the input images for the depth estimation network, but keeping RGB images for the image translation network, and (ii) augmenting the input images for both the depth estimation network and the image translation networks. Compared with the first augmentation strategy, the second strategy introduces a larger improvement. We believe it is because the semantic information can impose additional constraints on object geometry and these constraints are useful for regulating the shape of objects and determining the depth prediction on the boundary of objects. Therefore, applying semantic augmentation for the image translation network and the depth prediction network is expected to further boost the depth prediction accuracy. 

\subsubsection{Synthetic Translated Supervised + Real Self-Supervised}
By adding photometric losses to real-world sequential images and jointly training the synthetic supervised and self-supervised depth estimation, the depth estimation accuracy on real-world dataset is further improved by 11\%. 
The real sequential images are typically clear and accurate, which compensates for the shortcomings in the imperfect translated images. However, the photometric losses for real-world sequential images are computed based on an assumption that the displacement of pixels is purely caused by movement of the camera. Such an assumption does not always hold. The direct supervision on the synthetic translated images can help alleviate the negative effect of violating this assumption. But, our study indicates that by selecting valid pixels and filtering out the pixels that violates the assumption, our model is further improved by a noticeable margin. Finally, augmenting the input images with semantic labels yields even further improvement.

\begin{table}[hbt]
\begin{center}
\resizebox{\textwidth}{!}
{
\begin{tabular}{p{6cm} || c|c|c|c | c|c|c}
    \hline
    \multirow{2}{*}{Method}  
    & \multicolumn{4}{c|}{Error-related metrics} & \multicolumn{3}{c}{Accuracy-related metrics} \\
    & \textbf{Abs Rel} & \textbf{Sq Rel} & \textbf{RMSE} & \textbf{RMSE log}
    & $\mathbf{\delta < 1.25}$ & $\mathbf{\delta < 1.25^{2}}$  & $\mathbf{\delta < 1.25^{3}}$ \\
    \hline
\hline
\multicolumn{8}{c}{Synthetic Translated Supervised} \\ \hline
Without synthetic-to-real image translation & 0.278 & 3.216 & 6.268 & 0.322 & 0.681 & 0.854 & 0.929 \\ \hline
Native synthetic-to-real image translation & 0.168 & 1.199 & 4.674 & 0.243 & 0.772 & 0.912 & 0.966 \\ \hline
With flow-guided photometric consistency & 0.1539 & 0.993 & 4.4492 & 0.2241 & 0.7986 & 0.9356 & 0.9752
 \\ \hline 
With semantic consistency (frozen $S_{seg}$) &  0.1555 &    0.9680 &    4.7412 &    0.2324 &    0.7773 &    0.9245 &    0.9721
\\ \hline
With semantic consistency &  0.1544 &    0.9633 &    4.7422 &    0.2322 &    0.7786 &    0.9241 &    0.9727
\\ \hline
\hline
\multicolumn{8}{c}{Synthetic Translated Supervised with Semantic Augmentation} \\ \hline
Semantic augmentation for depth estimation network input only& 0.1532	& 0.9631 & 4.3872 & 0.2275 & 0.7945 & 0.9325 & 0.9738
 \\ \hline
Semantic augmentation for both the image translation networks \& the depth estimation networks& 0.1455	& 0.8869	& 4.2177	& 0.2154 & 0.8133 & 0.9411 & 0.9773
 \\ \hline
\hline
\multicolumn{8}{c}{Synthetic Translated Supervised + Real-world Self-Supervised} \\ \hline
With self-supervised depth estimation & 0.1292 & 0.6969 & 3.8399 & 0.1964 & 0.8428 & 0.9554 & 0.9826
 \\ \hline
With auto-masking & 0.1198 & 0.6671 & 3.7696 & 0.1921 & 0.8637 & 0.9583 & 0.9819 \\ \hline
With semantic augmentation & \textbf{0.1183} & \textbf{0.6150} & \textbf{3.7105} & \textbf{0.1876} & \textbf{0.8620} & \textbf{0.9622} & \textbf{0.9844}
\\ \hline
\end{tabular}
}
\end{center}
\caption{Performance gain in depth estimation from different model components. The predicted depth is capped at 50m}\label{tab:ablation_study}
\end{table}


\section{Conclusion and Next Steps}
In this paper, we present a framework for monocular depth estimation which combines the features of both synthetic images and real video frames in a novel semantic-aware, self-supervised setting. The complexity of our model does not affect its scalability, as we only require a depth network during inference time. We outperform all existing approaches on the KITTI benchmark as well as on our generalization to the new Make3D dataset. These factors contribute to the increased accuracy, scalability, and robustness of our framework as compared to other existing approaches. Our framework extends typical dataset-specific models to improve generalization performance, making it more relevant for real world applications. In the future, we plan to explore strategies which can apply similar frameworks to other related tasks in visual perception.

\clearpage

%
%
\bibliographystyle{splncs04}
\bibliography{reference}
\end{document}


\pagestyle{headings}
\mainmatter
\def\ECCVSubNumber{7036}  

\title{$S^3$Net: Semantic-Aware Self-supervised \\Depth Estimation with Monocular Videos and Synthetic Data \\
\ \\
\subtitle{Supplementary Material}
} 

\author[1]{Bin Cheng}
\author[2]{Inderjot Singh Saggu}
\author[3]{Raunak Shah}
\author[4]{Gaurav Bansal}
\author[5]{Dinesh Bharadia}

\affil[1]{Rutgers University, NJ, USA}
\affil[2]{Plus.ai, Cupertino, USA}
\affil[3]{Indian Institute of Technology, Kanpur, India}
\affil[4]{Blue River Technology, Sunnyvale, USA}
\affil[5]{University of California, San Diego, USA}
\institute{}
\maketitle

\begin{abstract}
In this supplementary material, we provide additional comparison results between our proposed $S^3$Net and two representatives of previous works, i.e., GeoNet~\footnote{GeoNet is a representative of approaches that predict depth using real videos.} and $T^2$Net~\footnote{$T^2$Net is a representative of the approaches utilizing synthetic depth ground-truth and image translation networks for depth prediction.}. We first present additional qualitative comparison of predicted depth maps generated by the three models. It can be noticed that the depth maps generated by our model posses higher visual quality, and suffer less from semantic distortion and object misdetection. We believe this is primarily because our model combines the merits of supervised depth prediction with domain adaption and self-supervised depth prediction. Secondly, we show that due to the semantic-aware design, the synthetic-to-real image translation module in our model can largely preserve the semantic structure during the image translation. Additional, the semantic-aware design also help improve depth predictions of our model for specific semantic categories and enhance the robustness of our model when the brightness of the input images varies. Lastly, we qualitatively compare the performance of the three models using a unseen dataset and demonstrate the better generalization capability of our model.
\end{abstract}

\section{Visualization Results for Depth Estimation}
\begin{figure}
    \centering
    \includegraphics[width=0.85\textwidth]{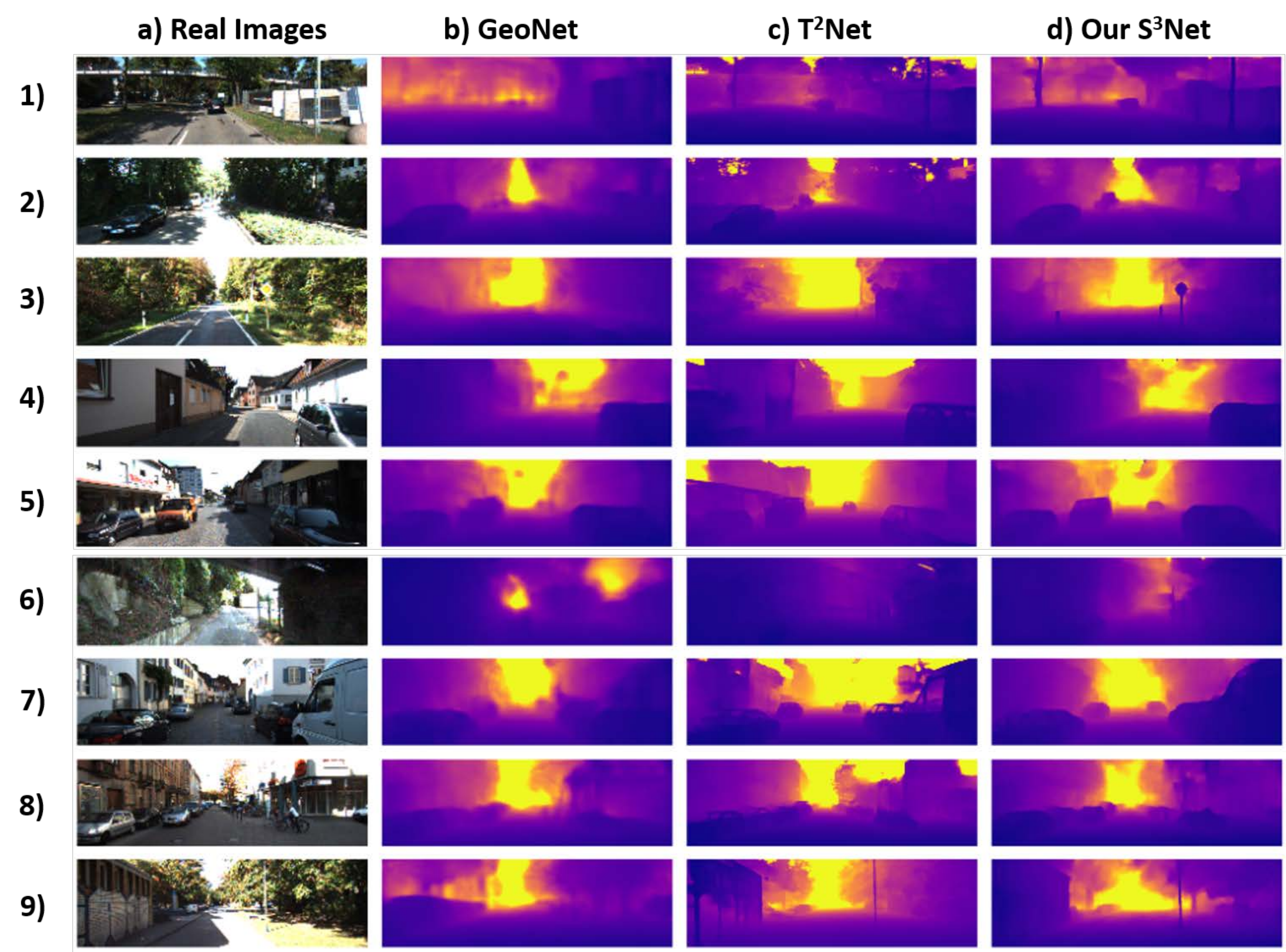}
    \caption{Additional predicted depth maps: $T^2$Net produces sharper depth maps than our model, but it fails to predict accurate depth for objects with texture changes. GeoNet doesn't suffer from these drawbacks but produces results which are blurred. Our approach generates sharper depth maps than GeoNet while being more accurate than both GeoNet and $T^2$Net in its predictions.}
    \label{fig:depth_extra}
\end{figure}

In this section we present additional depth estimation results (see Fig.~\ref{fig:depth_extra}) for the comparison between three models:1) GeoNet~\cite{geonet2018}; 2) $T^2$Net~\cite{t2net2018}; 3) our proposed  $S^3$Net.

\begin{figure}
    \centering
    \includegraphics[width=0.85\textwidth]{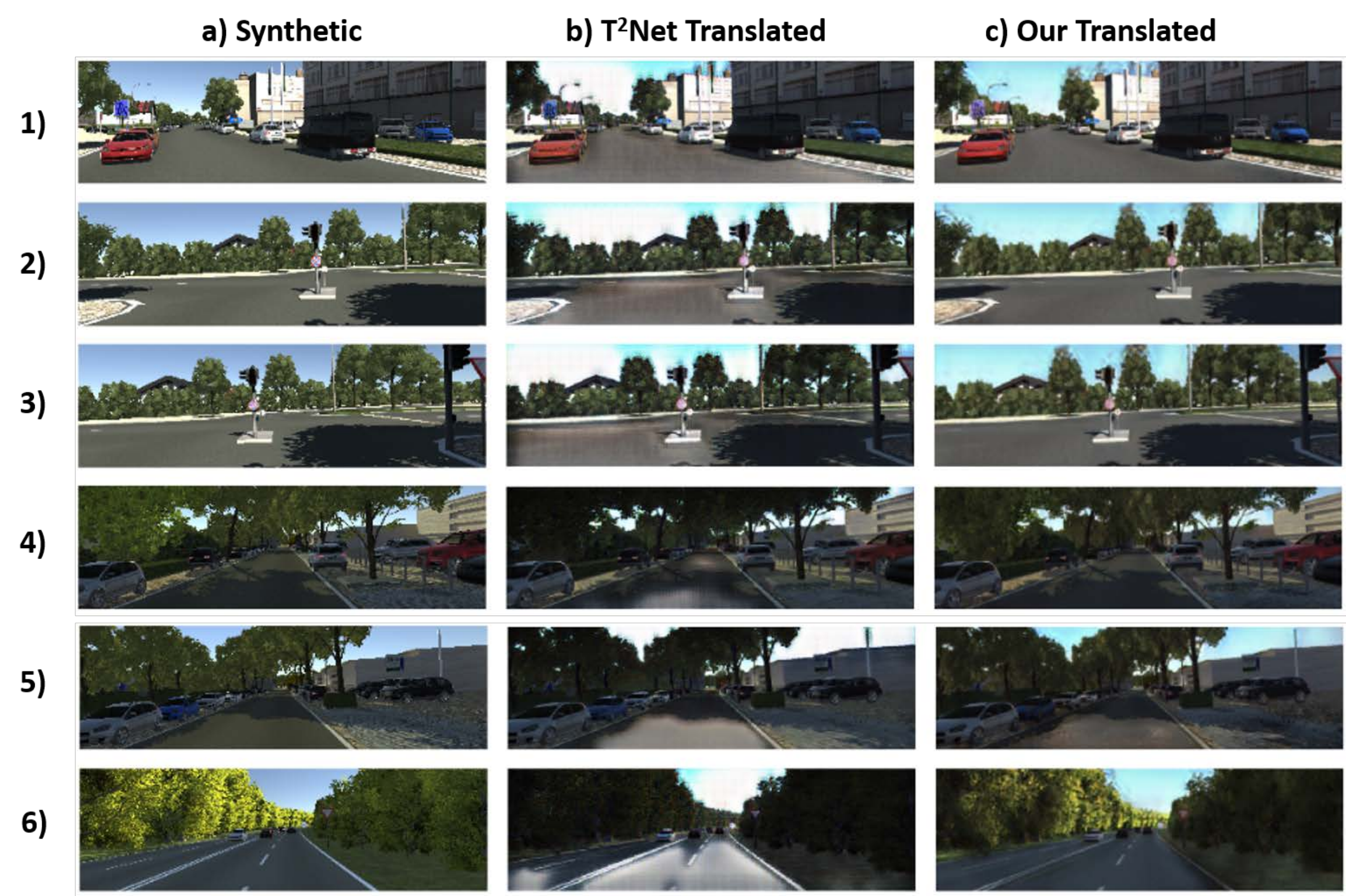}
    \caption{Normal to bright light conditions : Native $T^2$Net produces an artifact ridden translated image. Our semantic consistent translator significantly reduces the artifacts and retains the semantic structure of the image.}
    \label{fig:translated_a}
\end{figure}

Column b) in Fig.~\ref{fig:depth_extra} shows the predicted depth maps for GeoNet which is trained on real-world videos only in self-supervised fashion. GeoNet's results are not very sensitive to edge transitions and thus look blurry overall. $T^2$Net, on the other hand, can produce visually sharper depth maps but it fails to accurately predict depth for objects with changing texture, e.g., in image 1.c) different depth values are predicted for part of the same object over and under the bridge. This can also be observed in image 7.c) (house with window on the left) and image 9.c) (textured wall on the left) where drastic changes in depth are predicted for the same object due to change in texture.  
We also observe that $T^2$Net has troubles in distinguishing pedestrians and road signs from the background (image 2 and 3) whereas our $S^3$Net generates much better predictions. We also observe that for white shiny surfaces like in image 5 (building on the left) and in image 7 (building on the right)  $T^2$net predicts very high depth value. Looking at the translation results for $T^2$Net in Fig.~\ref{fig:translated_a} we can observe that the translated sky looks very very similar to the white shiny surfaces which can possibly explain the erroneous predictions.
Our predictions combine the best of both worlds, they are much sharper than GeoNet and don't have any of the issues discussed for $T^2$Net. However our model seems to have trouble in predicting depth for sky.

\section{Semantic Consistent Synthetic-to-Real Image Translation}
\begin{figure}
    \centering
    \includegraphics[width=0.85\textwidth]{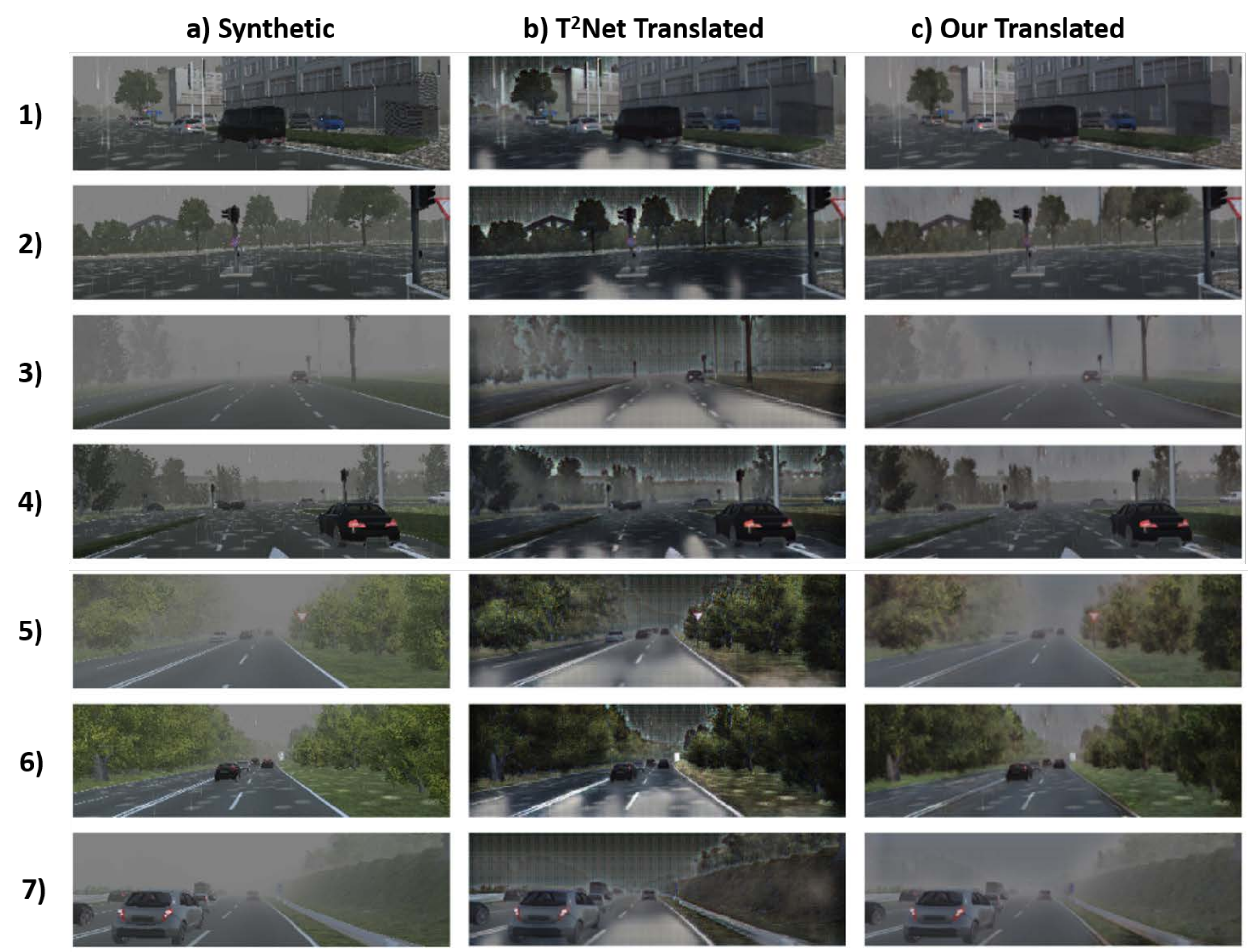}
    \caption{Low-light, fog and rainy conditions: $T^2$Net produces unsatisfactory translation results for synthetic images in low-light, foggy and/or rainy conditions. It produces checkered artifacts in the sky and severely distorts the road. Our additional semantic constraint allows translator network to perform well in these varying conditions.}
    \label{fig:translated_b}
\end{figure}

We present a qualitative comparison between native $T^2Net$ synthetic-to-real translator network versus our semantic consistent network in Fig.~\ref{fig:translated_a} and Fig.~\ref{fig:translated_b}.

\section{Category-Specific Depth Estimation}
\begin{figure}
    \centering
    \includegraphics[width=0.8\textwidth]{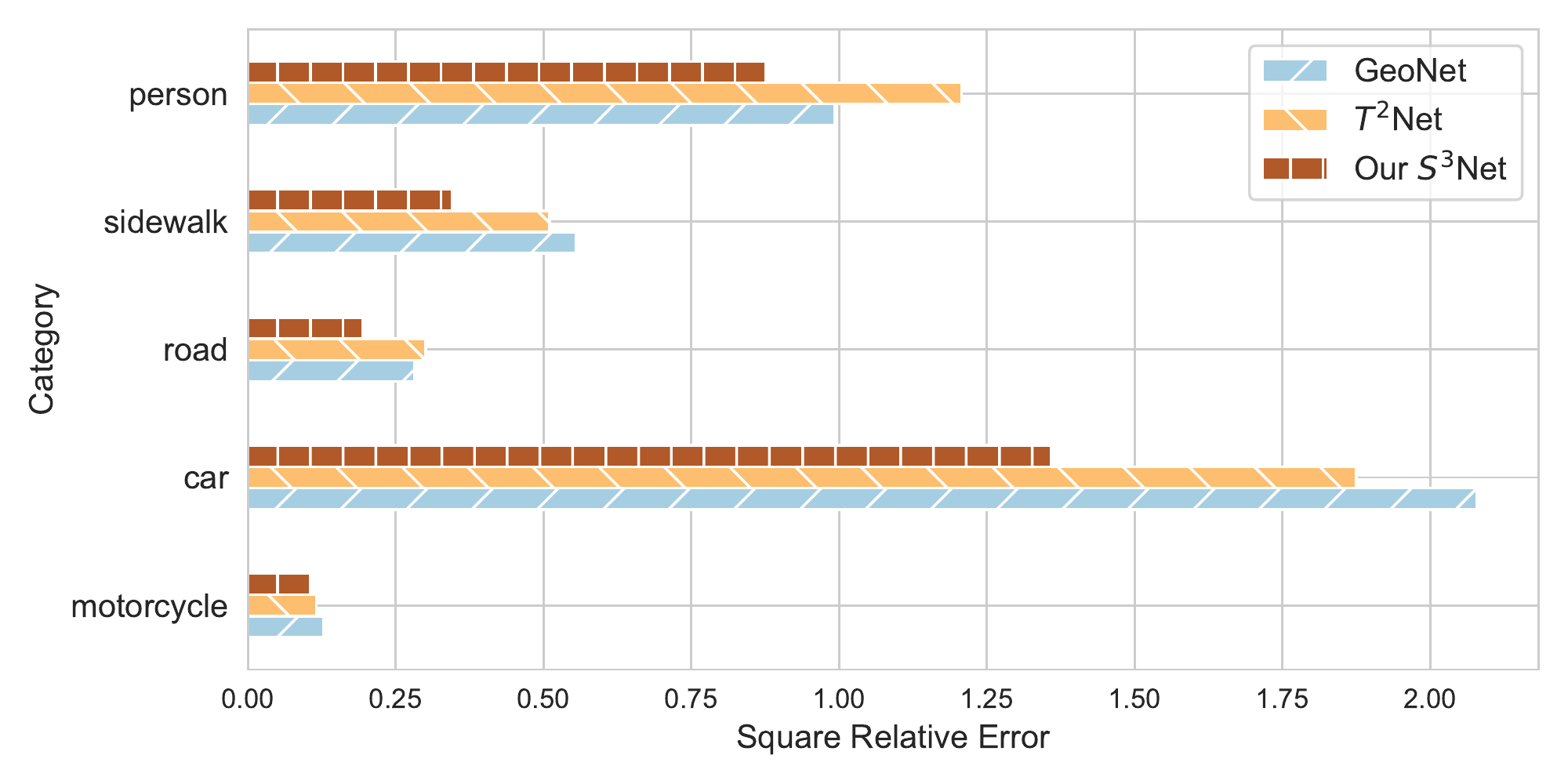}
    \caption{Squared relative error of predicted depths on KITTI dataset for various semantic categories}
    \label{fig:abs_rel_category}
\end{figure}

Fig.~\ref{fig:abs_rel_category} presents the depth prediction accuracy in terms of squared relative error for different semantic categories. As shown in the figure, our model outperforms GeoNet and $T^{2}$Net in all five selected categories. The improves are 34\%, 35\%, 37\%, 27\% and 13\% for categories \textit{car}, \textit{road}, \textit{sidewalk}, \textit{person} and \textit{motorcycle}, respectively. It can be noticed that the improvements for \textit{person} and \textit{motorcycle} are comparatively less, since these are overall less frequent categories in both synthetic and real data. We believe these improvements are primarily due to our model is semantic-aware and thus the semantic information is better preserved in depth prediction. 

\section{Depth Estimation Under Different Brightness Conditions}
\begin{figure}[h!]
    \centering
    \includegraphics[width=0.75\textwidth]{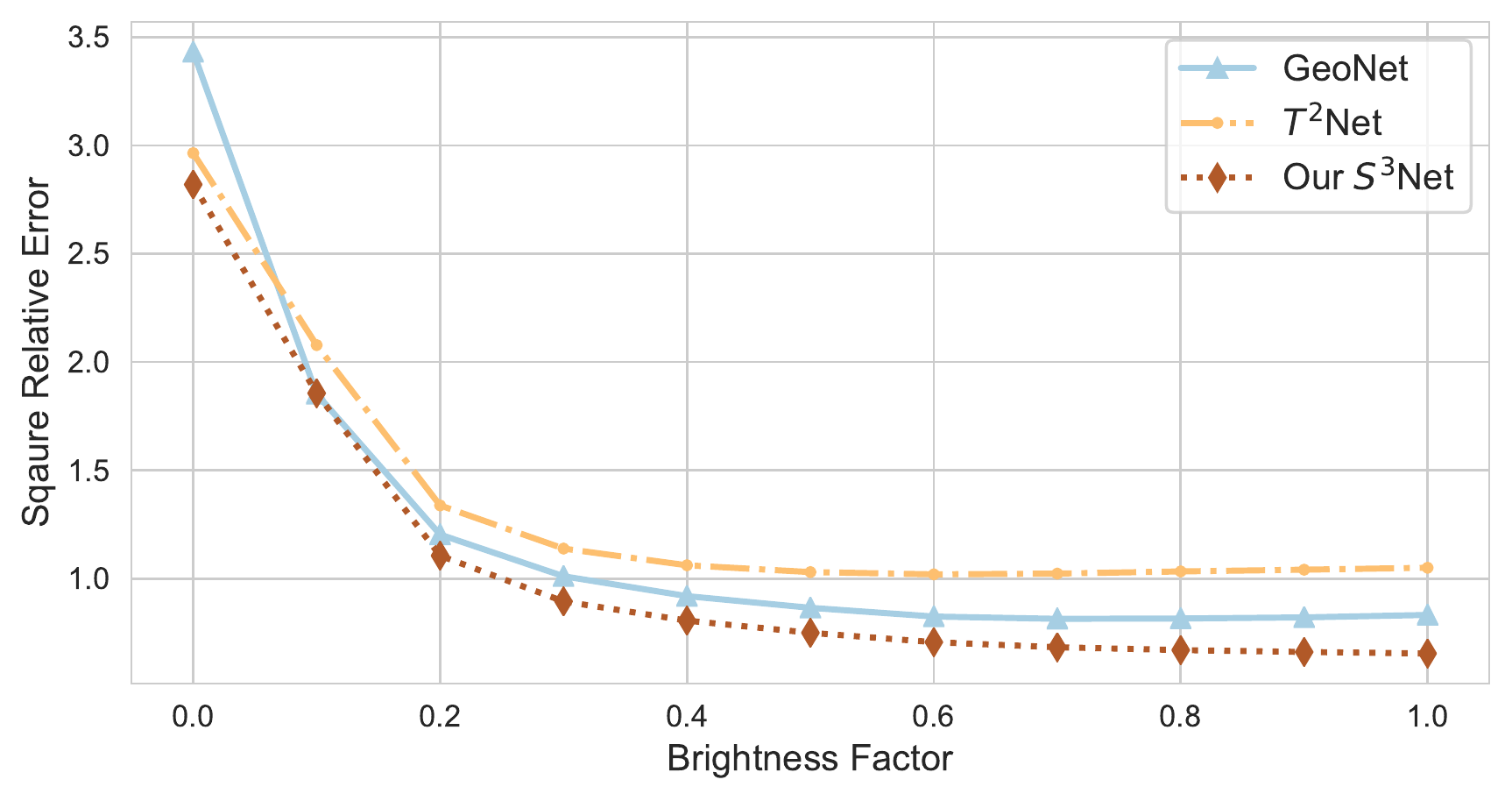}
    \caption{Squared relative error of predicted depths on KITTI dataset under various brightness conditions}
    \label{fig:abs_rel_brightness}
\end{figure}
We also compare the performance of different models for input images under various brightness conditions. Fig~\ref{fig:abs_rel_brightness} shows that our model consistently outperforms GeoNet and $T^2$Net(the larger the brightness factor is, the brighter the image is). We believe this is also because our model takes semantic information into consideration. The semantic information is robust under different brightness conditions, which can further enhance the robustness of our model.

\section{Qualitative Generalization on Make3D}
\begin{figure}
    \centering
    \includegraphics[width=0.85\textwidth]{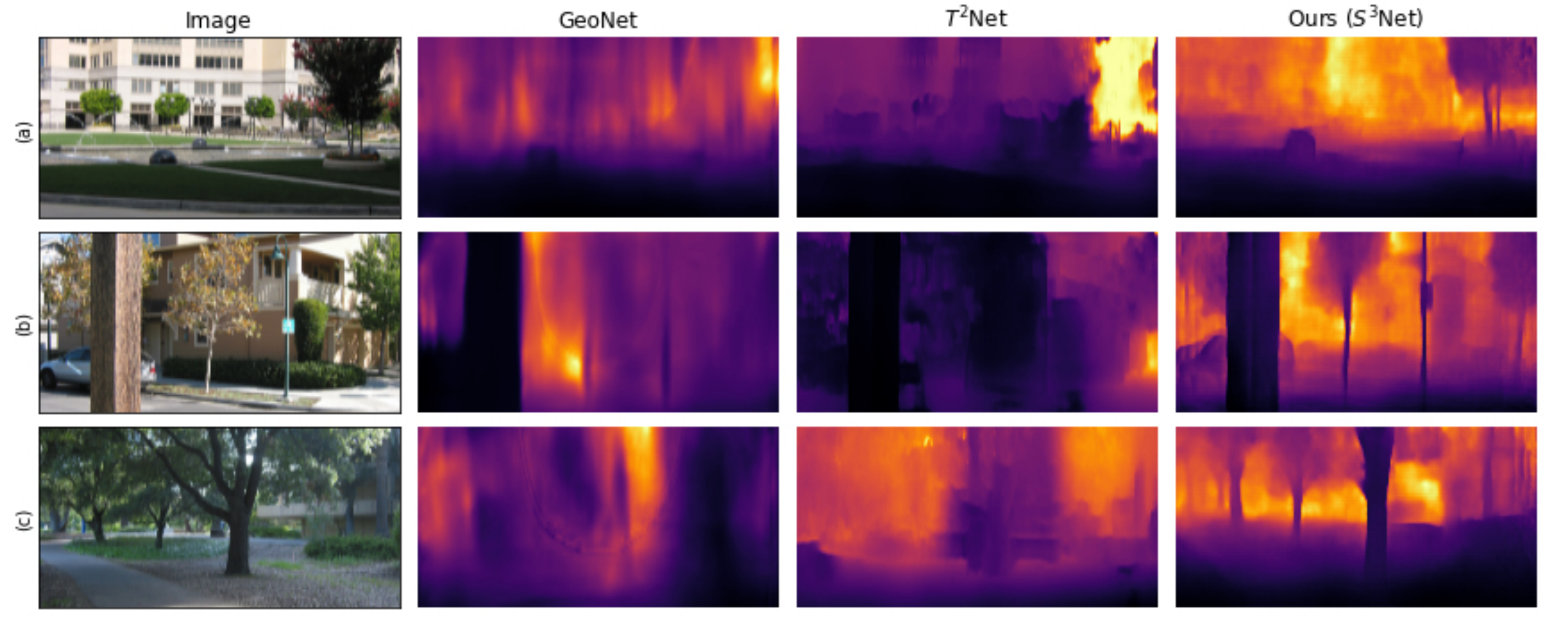}
    \caption{Examples of our $S^3$Net's generalization when as compared to a standard unsupervised model (GeoNet) and a standard syn to real model ($T^2$Net). Images taken from the Make3D test dataset.}
    \label{fig:make3d_eval}
\end{figure}

We present an example of our model's generalization capability through a qualitative comparison on the Make3D dataset, since this dataset is unseen and the images in this dataset are significantly different from the usual KITTI/vKITTI images. 
As shown in Figure~\ref{fig:make3d_eval}, our model can combine the merits of both supervised depth estimation with domain adaptation and self-supervised depth estimation. For example for input image (b), GeoNet partially succeeds in predicting the depths of the house and poles, whereas T2Net manages to understand variation of depths in intricate locations. Our model outperforms these two models in terms of both predicting visually more accurate depths of objects and clearly reflecting the depth changes in intricate locations.

\bibliographystyle{splncs04}
\bibliography{reference}